\documentclass[final,3p,times,authoryear]{elsarticle}

\usepackage{amssymb}
\usepackage{multirow}
\usepackage{amsmath}
\usepackage{subfig}
\usepackage{color}
\DeclareMathOperator*{\argmax}{argmax}
\usepackage{graphicx}

\begin{document}

\begin{frontmatter}
\title{An In-field Automatic Wheat Disease Diagnosis System}
\author[a,b]{Jiang Lu}
\ead{lu-j13@mails.tsinghua.edu.cn}
\author[a]{Jie Hu}
\ead{huj14@mails.tsinghua.edu.cn}
\author[a]{Guannan Zhao}
\ead{zgn14@mails.tsinghua.edu.cn}
\author[b]{Fenghua Mei}
\ead{mei\_fh@21cn.com}
\author[a]{Changshui Zhang\corref{cor1}}
\ead{zcs@mail.tsinghua.edu.cn}
\cortext[cor1]{Correspoding author.}
\address[a]{Department of Automation, Tsinghua University, 
State Key Lab of Intelligent Technologies and Systems, Tsinghua National Laboratory for Information Science and Technology (TNList), Beijing, PR China}
\address[b]{China Marine Development and Research Center (CMDRC), Beijing, PR China}

\begin{abstract}
Crop diseases are responsible for the major production reduction and economic losses in agricultural industry worldwide. Monitoring for health status of crops is critical to control the spread of diseases and implement effective management. This paper presents an in-field automatic wheat disease diagnosis system based on a weakly supervised deep learning framework, \emph{i.e.} deep multiple instance learning, which achieves an integration of identification for wheat diseases and localization for disease areas with only image-level annotation for training images in wild conditions. Furthermore, a new in-field image dataset for wheat disease, \emph{Wheat Disease Database 2017} (WDD2017), is collected to verify the effectiveness of our system. Under two different architectures, \emph{i.e.} VGG-FCN-VD16 and VGG-FCN-S, our system achieves the mean recognition accuracies of 97.95\% and 95.12\% respectively over 5-fold cross-validation on WDD2017, exceeding the results of 93.27\% and 73.00\% by two conventional CNN frameworks, \emph{i.e.} VGG-CNN-VD16 and VGG-CNN-S. Experimental results demonstrate that the proposed system outperforms conventional CNN architectures on recognition accuracy under the same amount of parameters, meanwhile maintaining accurate localization for corresponding disease areas. Moreover, the proposed system has been packed into a real-time mobile app to provide support for agricultural disease diagnosis.
\end{abstract}

\end{frontmatter}

\section{Introduction}
\label{sec1}
Crop disease diagnosis is of great significance to prevent the spread of diseases and maintain the sustainable development of agricultural economy. In general, the crop disease diagnosis is performed manually by visual observation or microscope techniques, which are proven to be time-consuming and have the risk of error due to subjective perception. In this context, various spectroscopic and imaging techniques have been studied for identifying crop disease symptoms \citep{bravo2004foliar,moshou2005plant,chaerle2007multicolor,belasque2008detection,qin2009detection}. Although these techniques can make a relatively rapid diagnosis for crop diseases, they can't be separated from the support of expensive and bulky sensors.

\begin{figure}[htbp] 
\centering
\includegraphics[width=0.8\linewidth]{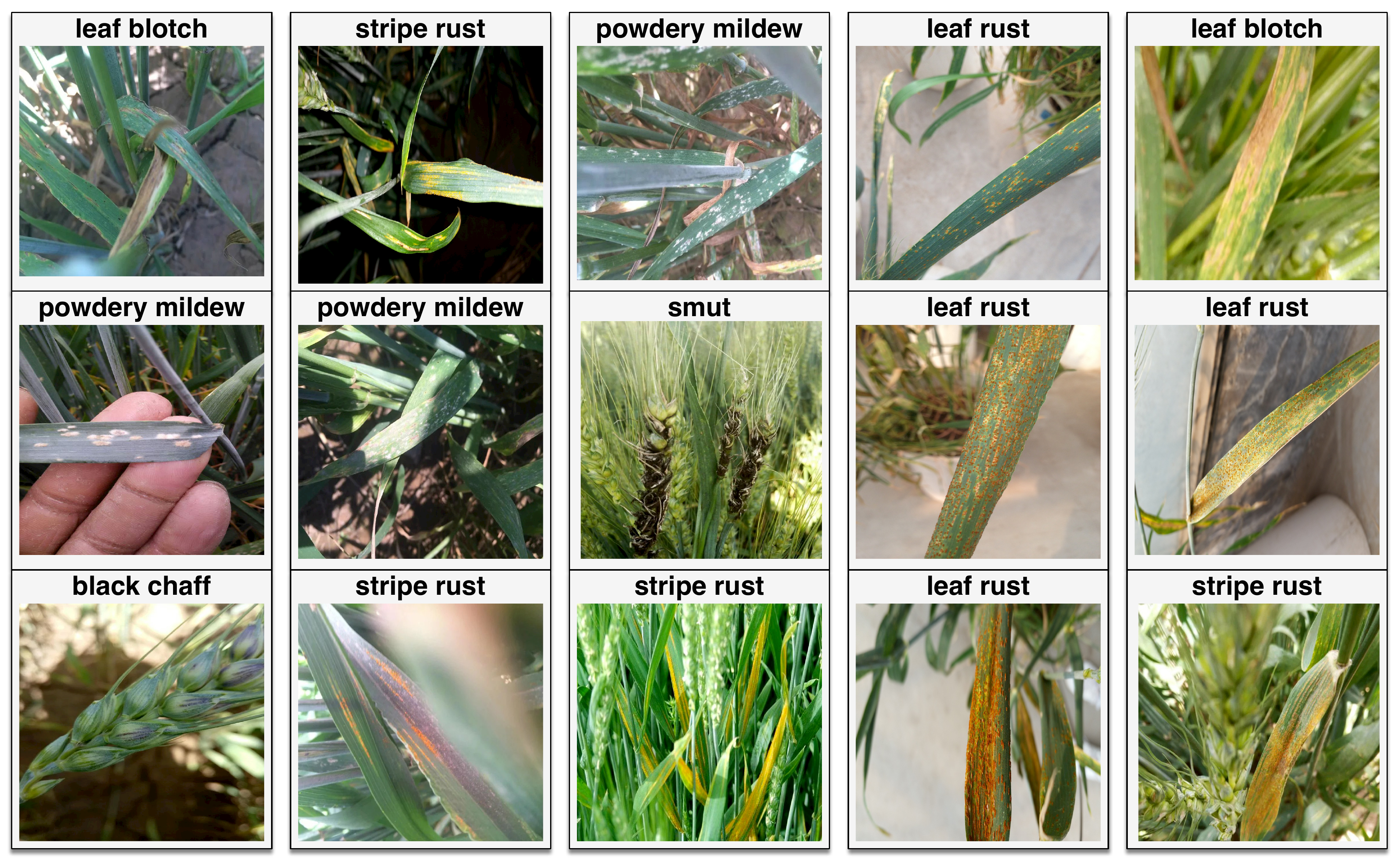}
 \caption{Samples selected from our collected dataset WDD2017. These images are captured under in-field environments. Five columns from left to right shows the five intractable challengens to in-field wheat disease diagnosis as mentioned in Section~\ref{sec1}. The images in first colunm contains complex backgrounds. The images in second column shows different capture conditons, such as illumination changes, image blur caused by camera shakes. The third column demonstrates the coexistence of multiple leaves or multiple disease areas in one image. The fourth column and fifth column reveal the various characterization for different period of{} disease development and the similar appearance between different wheat diseases respectively.}
 \label{WDDpics}  
\end{figure}

With the development of computer vision, concerns are growing about the image-based detection technologies for crop disease \citep{camargo2009image,arivazhagan2013detection,barbedo2014automatic,rastogi2015leaf}, which gets rid of the shackles of time cost and molecular analysis equipments \citep{martinelli2015advanced}. Instead, people just need common cameras and consumer-level electronic storage devices to perform crop disease identification. However, these methods are almost task-specific and need expert knowledge to design handcrafted feature extractors, only working for the crop images under ideal experimental environment. To make an automatic crop disease diagnosis system that can be applied to in-field images, one has to face some intractable challenges summarized in \citep{barbedo2016review}: (1) complex image backgrounds, \emph{e.g.} leaves, soils, stones and even the people's hands, (2) uncontrollable capture conditions, \emph{e.g.} illumination, camera angle and image quality, (3) co-occurrence of multiple leaves or multiple disease areas in one image, (4) various characterization for different stages of disease development, (5) similarities in appearance between different disease categories. Some challenging samples, which are selected from our collected dataset of wheat diseases, are demonstrated in Figure \ref{WDDpics}. To our best knowledge, few researches have been done to mitigate or eliminate the above challenges for crop disease diagnosis.

The overall objective of this work is to develop an automatic wheat disease diagnosis system to identify disease categories and locate corresponding disease areas simultaneously for in-field wheat images. In order to avoid the expensive and laborious manual annotation, the twofold task of identification and localization for wheat diseases is modeled as a weakly supervised learning task. As far as we know, we firstly propose to jointly handle the twofold task for wheat diseases in an in-field scenario.

\begin{figure*}[!htbp] 
\centering
\includegraphics[width=1.\linewidth]{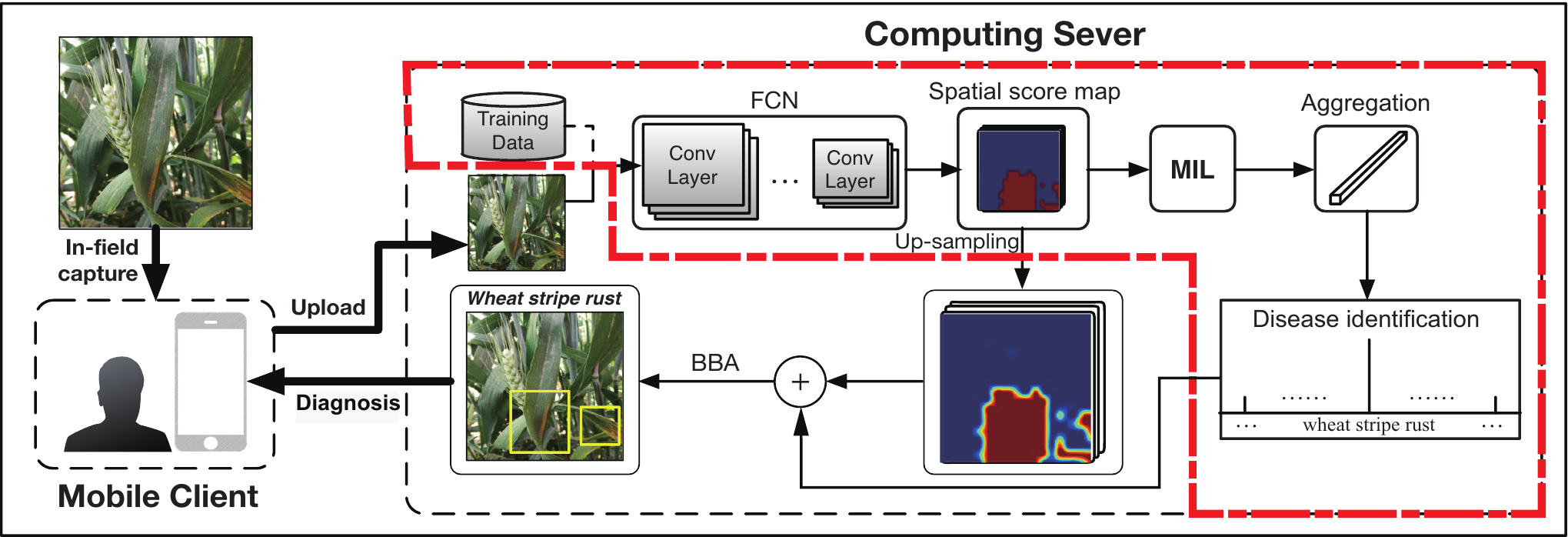} 
 \caption{Technical pipeline of our DMIL-WDDS. The parts in left and right dotted boxes
 are mobile client and computing sever respectively. The components surrounded by red bold dotted line will go through a training stage on training data before this system performs disease identification and localization. The diagnoses to be fed back to users consist of disease category and positions of corresponding disease areas. Best viewed in color.}
 \label{pipeline}  
\end{figure*}

In this paper, an novel in-field automatic wheat disease diagnosis system is proposed based on deep learning and  multiple instance learning (MIL) \citep{dietterich1997solving}, which can be deployed on mobile handsets to perform real-time diagnosis. As showed in Figure \ref{pipeline}, our framework is composed of mobile client and computing sever in device level. On the one hand, a fully convolutional network (FCN) is exploited to perform local feature extraction and local disease estimation for a resized image captured from mobile camera in wild conditions. Consequently, the FCN produces some spatial score maps over different disease species where each score point corresponds to a particular respective local window of raw image. Then these estimations of different local windows are fed into MIL framework to aggregate the overall evaluation for the whole image. On the other hand, the spatial score maps go through an up-sampling operation to roughly delineate the disease location, then a bounding boxes approximation (BBA) step is performed to accurately lock disease positions. The proposed model is named multiple instance learning based wheat disease diagnosis system (DMIL-WDDS). In particular, the DMIL-WDDS can achieve end-to-end training with enough prepared training data. To verify the practicability and effectiveness of our DMIL-WDDS, an in-field disease dataset \emph{Wheat Disease Database 2017} (WDD2017) is collected, which consists of 9,230 images with 7 different classes (6 for common wheat diseases, 1 for healthy wheat). The results of experiments on WDD2017 show that the proposed DMIL-WDDS outperforms conventional CNN architectures on recognition accuracy for disease categories, but also keeping precise localization for corresponding disease areas. 

Our main contributions are summarized as:

$\bullet$ A weakly supervised learning framework based on DMIL is firstly exploited for wheat disease diagnosis, which has the ability to cope with intractable in-field wheat images.

$\bullet$ An integration of identification and localization for wheat diseases is achieved by proposed system, which outperforms conventional CNN-based recognition architecture under the same amount of parameters in deep model.

$\bullet$ A new in-field wheat disease dataset WDD2017 is collected to demonstrate the effectiveness of proposed system as well as building a benchmark for subsequent works.

\section{Related work}
\label{sec2}
Many image-based methods have been developed to handle crop disease identification. Based on the fusion of hyper-spectral and multi-spectral fluorescence measurements, \citep{moshou2005plant} presented a self-organizing map (SOM) based disease classifier. \citep{camargo2009image} proposed to identify the visual symptoms of crop diseases via color transformation and image color segmentation. In \citep{phadikar2013rice}, Fermi energy based segmentation method was proposed to isolate infected region from whole image, then rough set theory (RST) was used to feature selection and rule base classifier was used to disease identification. In brief, the majority of existing methodologies have the limitations of pure background or controlled environment for captured images, which is a dilemma for practical application. As a result, leaf segmentation \citep{zhang2011automatic,alenya2013robotized,wang2013adaptive} was  usually compelled to the prerequisite for performing the crop disease identification. However, one will fall into another dilemma by naive leaf segmentation as handling images where multiple leaves or multiple infected regions coexist. Although recent works about computer vision proved the great success of \emph{selective search} (SS) \citep{uijlings2013selective} in object segmentation, it seems impotent to directly adopt SS to segment the desultory in-field images due to the serious overlapping and crisscross between crops.

Deep convolutional neural network (DCNN) has accelerated the development of computer vision to some extent \citep{krizhevsky2012imagenet,simonyan2014very,szegedy2015going,sermanet2013overfeat,girshick2014rich}. Given large-scale training data and high-performance computing infrastructure, DCNN can lead to striking performance in classification and object detection. Of course, DCNN has been applied in the field of precision agriculture in recent years. \citep{dyrmann2016plant,grinblat2016deep} designed their task-specific DCNN architectures to perform plant species classification. \citep{sa2016deepfruits} proposed a multi-modal Faster R-CNN model, namely \emph{DeepFruits}, to play real-time fruit detection. In addition, a small amount of researches focused on plant disease detection using DCNN. \citep{sladojevic2016deep} collected 3,000 original leaf images from Internet and performed data augmentation process on this database, then built a DCNN to automatically classify and detect 13 different types of plant diseases from leaf images. Similarly, \citep{mohanty2016using} trained a DCNN on a dataset of 54,306 images of diseased and healthy plant leaves to recognize 14 crop species and 26 diseases. However, the above deep learning approaches for plant disease detection  built their deep recognition models on pure images collected under controlled conditions, which are not applicable to the wild environment. Besides, they just realized the identification of diseases but paid no attention to where the diseases are.

MIL was firstly introduced in \citep{dietterich1997solving} for drug activity prediction in a weakly labeled scenario. To reduce the laborious manual annotation as much as possible and achieve object localization simultaneously, some works are focused on the combination of deep learning and MIL framework \citep{pinheiro2015image,wu2015deep}. \citep{wu2015deep} made an assumption for MIL that the desired objects must lie among all region proposals which can be produced via off-the-shelf region proposal algorithms, \emph{e.g.} SS. Because of the computing cost caused by region proposal algorithms and the complexity of leaves distribution, the fashionable region segmentation or proposal methods may step into a trouble for practicality. 

FCN was firstly proposed in \citep{long2015fully} for semantic segmentation. In this paper, a new FCN is exploited to reduce computing cost and perform instance-level disease estimation simultaneously, which equates to a sliding window operation on the whole image.

In brief, little researches concentrate on crop disease identification and localization upon in-field raw images in complex cluttered scenes, and no available dataset for in-field crop disease images has come out up to now. On the one hand, our system aims at in-field automatic wheat disease diagnosis and is more proximate to the practical situations, which will provide technical supports and services for precision agriculture. On the other hand, the WDD2017 is the first proposed in-field crop diseases dataset, which will build a benchmark on in-field disease detection and promote subsequent relevant works. Hence, our work is valuable for further studies.

\section{Materials and methods}
\label{sec3}

\subsection{Wheat Disease Database 2017 (WDD2017)}
\label{sec3.1}
To facilitate the researches of crop disease diagnosis, some relevant datasets have been released openly and freely, \emph{e.g.} \emph{PlantVillage}, which includes over 50,000 expertly
annotated images on healthy and infected leaves of crops. However, it is a pity that all images in PlantVillage have been processed into an ideal state which is hard to be seen in wild conditions, namely one crop leaf in a pure background. To our best knowledge, no suitable dataset for in-field crop disease diagnosis is available so far. Accordingly, some human and material resources are devoted to collect 9230 wheat crop images with 7 different wheat diseases including healthy class, which are annotated in image level by agricultural experts. This dataset of collected images is named as Wheat Disease Database 2017 (WDD2017), and some examples of WDD2017 are illustrated in Figure~\ref{WDDpics}. Note that WDD2017 has been uniformly divided into 5 folds to play cross-validation, where 4 folds are used as training set and 1-fold is used as test set. Table \ref{Tab1} shows the details of our WDD2017.  

\begin{table}[htbp]
\centering  
\caption{The composition of Wheat Disease Database 2017 (WDD2017).}
\vspace{-.5em}
\scalebox{0.95}{
\begin{tabular}{lll}  
\hline
Wheat Diseases & \# Images &  \# Train / Test Split \\ \hline
Powdery Mildew & 350 &280 / 70 \\
Smut & 1455 &1164 / 291\\
Black Chaff & 585 &468 / 117\\
Stripe Rust & 1755& 1404 / 351\\
Leaf Blotch & 2455&1964 / 491\\
Leaf Rust & 1110&888 / 222\\
Healthy Wheat & 1520& 1216 / 304\\ \hline          
\end{tabular}}
\label{Tab1}
\end{table}

The 6 kinds of diseases in Table \ref{Tab1} are common to wheat crops, which are responsible for the reducing yield of wheat crops in agricultural economy. To be clear, our WDD2017 has the following characteristics: (1) every image in WDD2017 almost only contains one kind of disease as which this image was annotated, (2) every image is collected under in-field scenario followed by no technical means, maintaining all primitive information of capture environment, (3) WDD2017 covers most of current challenges for wheat disease diagnosis as illustrated in Figure \ref{WDDpics}, including complex backgrounds, different capture conditions, various characterization for different stage of disease development (early, middle and late stage of diseases) and similar appearance between different wheat diseases.

\subsection{Aggregation by multiple instance learning (MIL)}
\label{sec3.2}
Multiple instance learning (MIL) is a weakly supervised learning methodology, which aims to decrease the labeling effort by only using annotations assigned to bags instead of instances. In MIL setting, the bag for one class containing at least one positive instance is labelled \emph{positive}, while the bag whose instances are all negative is labelled \emph{negative}. For the classification task, suppose $\mathbf{B}_{k}$ is a bag from the bags set $\{\mathbf{B}_{k}, k=1,\ldots,N\}$, and the instances of $\mathbf{B}_{k}$ are denoted as $\{B_{k1},\ldots,B_{kn_{k}}\}$, where $n_{k}$ denotes the number of instances for bag $\mathbf{B}_{k}$. $c_{k} \in \{1,\ldots,C\}$ is the class label for bag $\mathbf{B}_{k}$, and $c_{kj} \in \{1,\ldots,C\}$ is the class label for instance $B_{kj}$. The conditional probability that instance $B_{kj}$ is class $c$ is denoted as $p_{kj}^{c}=P(c_{kj}=c| B_{kj})$, $c\in \{1,\ldots,C\}$. The next job is to find an aggregated function $\mathcal{F}$ that can suitably encode MIL assumption. More specifically, if an instance is likely to be positive for one class, the function $\mathcal{F}$ should assign more positive trusts to its corresponding bag. The aggregated function $\mathcal{F}$ for bag $\mathbf{B}_{k}$ can be expressed as:
\begin{equation}
p_{k}^{c}=P(c_{k}=c|\mathbf{B}_{k})=\mathcal{F}(p_{k1}^{c},p_{k2}^{c},\ldots,p_{kn_{k}}^{c})\quad,
\label{aggrefunc1}
\end{equation}  
where $\mathcal{F}$ can be $\mathrm{Max}_{j}(p_{kj}^{c})$, $\mathrm{Avg}_{j}(p_{kj}^{c})$ or others \citep{wu2015deep}. In our work, we also consider the Softmax aggregated function proposed in \citep{ray2005supervised}:
\begin{equation}
p_{k}^{c}=\mathcal{F}_s(p_{k1}^{c},p_{k2}^{c},\ldots,p_{kn_{k}}^{c})
=\frac{\sum_{j=1}^{n_{k}} p_{kj}^{c} \cdot e^{\alpha p_{kj}^{c}}}{\sum_{j=1}^{n_{k}}e^{\alpha p_{kj}^{c}}}\quad,
\label{aggrefunc2}
\end{equation}  
where $\alpha$ is a constant that control the extent to which the Softmax aggregated function approximates a hard max aggregated function. Essentially, this Softmax form makes a bag-level estimation  based on a weighted consideration on all instances in a bag. Note that we just rely on the bag-level label to perform the gradient-based back propagation algorithm in MIL assumption. 

\subsection{Fully convolutional network (FCN)}
\label{sec3.3}
\begin{figure}[tbp] 
\centering
\includegraphics[width=0.7\linewidth]{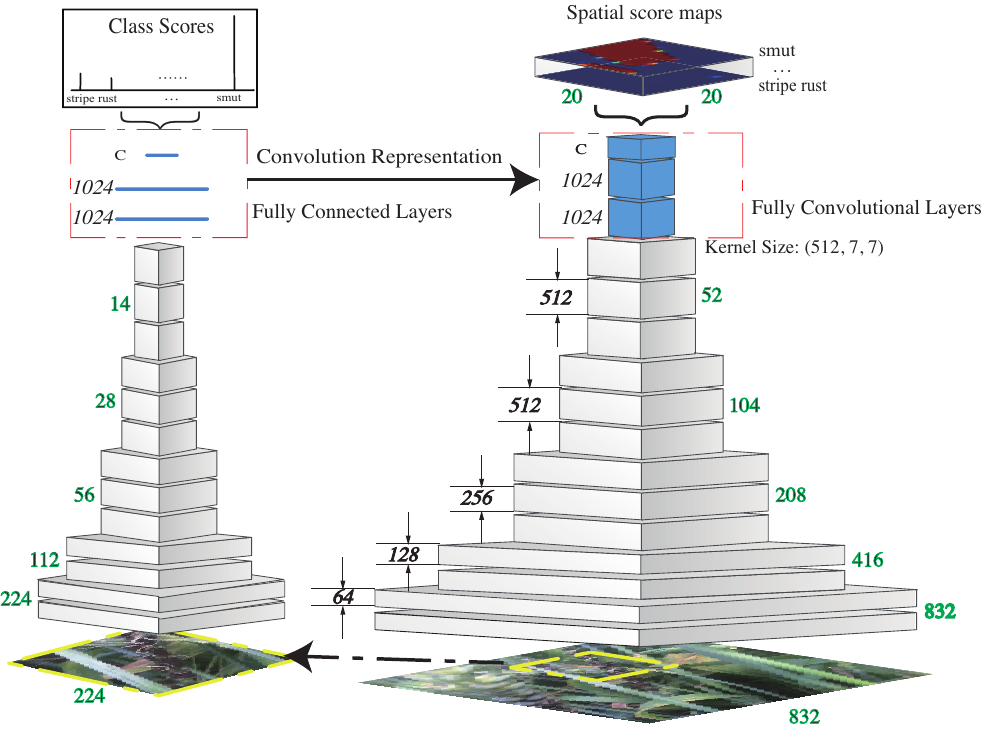} 
\vspace{-0.8em}
 \caption{Diagram of the transformation from modified VGG-CNN-VD16 to VGG-FCN-VD16. The numbers highlighted in green indicate the size of nearby feature maps or images, and the black italic numbers indicate the number of feature maps. Best viewed in color.}
 \label{FCN}  
\end{figure}

In general, the CNN consists of 4 parts: \emph{convolution layers}, \emph{pooling layers}, \emph{fully connected layers} and \emph{output layer}. CNN takes an image or a patch of an image as an input and outputs a probability distribution over all classes. The output of each convolution layer in CNN is a 3-dimensional tensor with size of $d\times h \times w$, called feature maps, where $h$ and $w$ are spatial dimensions of feature map, and $d$ is the number of feature maps or channels. Regions in higher layers correspond to particular regions in input image, which are referred to \emph{respective field}. Take VGG-CNN-VD16 for example \citep{simonyan2014very}, the respective field of its final output is actually the whole $224 \times 224$ image. Based on a tensor deformation, the fully connected layers can be transformed into convolution representations, as illustrated in Figure \ref{FCN}. In this work, the channels of fully connected layers for VGG-CNN-VD16 are modified from $\{4096,4096,1000\}$ to $\{1024,1024,C\}$ respectively, where $C$ is the number of classes in our task, therefore the size of weight matrix for first fully connected layer is $20588 \times 1024$. For convolution representation, the weights of first fully connected layer are reshaped into a 4-dimensional tensor with size of $1024\times 512 \times 7 \times 7$, which serves as 1024 convolutional kernels with size of $512 \times 7 \times 7$. By the same token, other fully connected layers are handled to obtain a FCN called VGG-FCN-VD16. Comparatively, a relatively shallow CNN architecture VGG-CNN-S \citep{chatfield2014return} is also introduced to play the same operation as VGG-CNN-VD16, forming a new FCN named VGG-FCN-S. 

Take VGG-FCN-VD16 for example, we can obtain a spatial score map for each disease as playing VGG-FCN-VD16 upon one raw image whose size exceeds $224 \times 224$. Intuitively, each score point in the map is approximately equivalent to the estimation generated by applying the original CNN to fixed-size sliding window over raw image. This architecture allow us to skim through all likely locations for striking objects. Considering that the disease areas of one image may occupy in small regions, we resize the raw image into the size of $832\times 832$, which leads to some $20 \times 20$ score maps under VGG-CNN-VD16 configuration. Explicitly, each score point corresponds to a $224 \times 224$ square area in the resized input image. Because of the five pooling layers and $2\times 2$ pooling size of VGG-CNN-VD16, the stride of sliding window is $2^{5}=32$. Compared with original CNN acting on sliding windows in raw image successively, VGG-FCN-VD16 achieve one-time calculation for all sliding patches, highly amortizing the computation over overlapping regions of those sliding windows and meeting the requirement of real-time application better.

\subsection{Bounding boxes approximation (BBA)}
\label{sec3.4}
\begin{figure}[tbp] 
\centering
\includegraphics[width=0.8\linewidth]{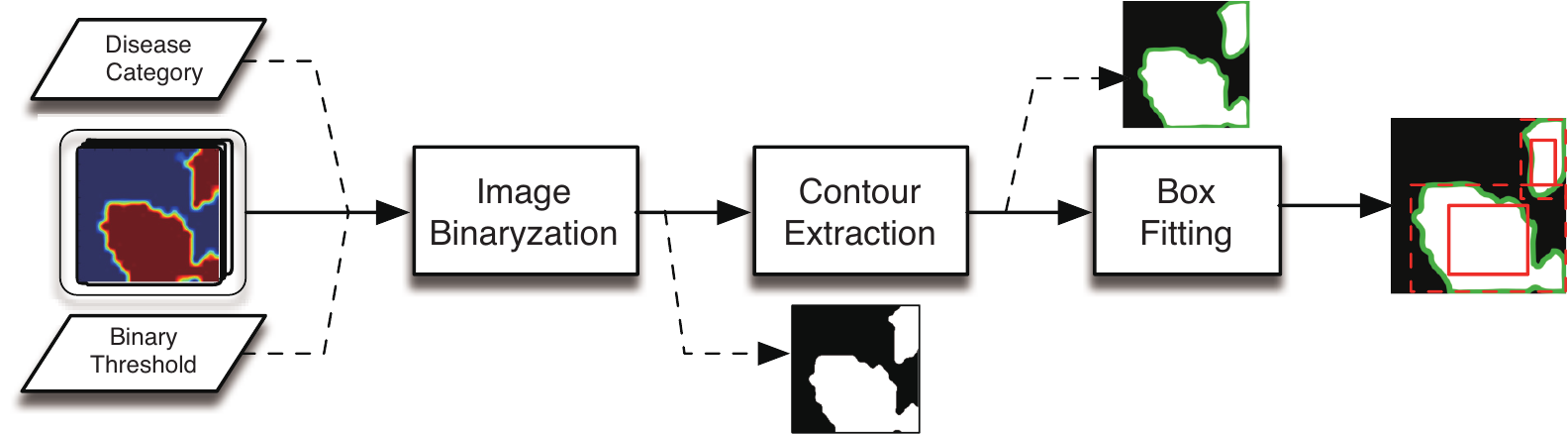} 
\vspace{-0.5em}
 \caption{Pipeline of bounding boxes approximation (BBA). The green line denotes the detected contour on binary image. The red dashed box denotes the smallest box containing the corresponding contour, while the red solid box denotes the scaled-down version of the box. Best viewed in color.}
 \label{BBA}  
\end{figure}

For clarity, a bounding boxes approximation (BBA) step is adopted to locate the position of disease areas. As showed in Figure \ref{BBA}, given the spatial score maps and corresponding disease category predicted by MIL framework, we can transform the spatial score map related to predicted disease channel into a binary image conditioned on a binary threshold, then a contour extraction operation is acted upon it. Afterwards, the smallest box that contains the corresponding contour can be generated. Taking into account the overlapping regions caused by FCN, we scale down the smallest box uniformly in certain proportion, which contributes to precise localization for disease areas. Fortunately, it is an easy job that using openCV\footnote{OpenCV (Open Source Computer Vision) is a library of programming functions mainly aimed at real-time computer vision.} to achieve all of the above processing steps.

\subsection{DMIL-WDDS}
\label{sec3.5}
As described earlier, the FCN is used for extracting local features from the whole image and generating spatial score maps where each individual score point is essentially a disease estimation for corresponding receptive field. Remarkably, FCN acts approximately as a sliding window operation, leading to the fact that the striking objects lie in at least one of the these receptive fields. In this sense, we treat the image whose receptive fields cover striking objects as a positive bag for its corresponding class label in MIL. As illustrated in Figure \ref{pipeline}, we divide our DMIL-WDDS model into two parts, one of which implements image-level classification, \emph{i.e.} wheat disease identification, and another implements disease area localization. The classification part, surrounded by red bold dotted line in Figure \ref{pipeline}, will pass a training stage to build a generalized deep model in advance. Once well trained, the deep model will be fixed to perform feature extraction and spatial score map generation for new images. 

To explain, we adopt ReLu as the nonlinear activation function of our networks to accelerate convergence, except for the last convolutional layer that generates spatial score maps. In order to get a probability value between 0 and 1, we use sigmoid function instead of ReLu in last fully convolutional layer. Given a new image $\mathbf{B}_{k}$, we denote by $h_{k}$ and $w_{k}$ the height and width of its spatial score maps respectively, and denote by $p^{c}_{k-ij}$ the output of sigmoid function for the location $(i,j)$ of spatial score map in channel $c$, \emph{i.e.} the probability that local receptive field $B_{k-ij}$ is classified as disease $c$, where $i = \{1,\ldots,h_{k}\}$, $j \in \{1,\ldots,w_{k}\}$ and $c \in \{1,\ldots,C\}$. Under MIL framework, if we use Softmax as our aggregated function, the probability of containing object $c$ by $\mathbf{B}_{k}$ can be wrote as:
\begin{equation}
p_{k}^{c}=\mathcal{F}_s(p_{k-11}^{c},p_{k-12}^{c},\ldots,p_{k-h_{k}w_{k}}^{c})
=\frac{\sum_{i=1}^{h_{k}} \sum_{j=1}^{w_{k}}  p_{k-ij}^{c} \cdot e^{\alpha p_{k-ij}^{c}} }
{\sum_{i=1}^{h_{k}} \sum_{j=1}^{w_{k}} e^{\alpha p_{k-ij}^{c}}}\quad,
\label{aggrefunc3}
\end{equation}  
where $\alpha$ is the same argument as Equation~(\ref{aggrefunc2}). Also, we can use $\mathrm{Max}$ aggregated function:
\begin{equation}
p_{k}^{c}=\mathcal{F}_m(p_{k-11}^{c},p_{k-12}^{c},\ldots,p_{k-h_{k}w_{k}}^{c})
=\mathrm{Max}_{i,j}(p_{k-ij}^{c})\quad,
\label{aggrefunc4}
\end{equation}  
or $\mathrm{Avg}$ aggregated function:
\begin{equation}
p_{k}^{c}=\mathcal{F}_a(p_{k-11}^{c},p_{k-12}^{c},\ldots,p_{k-h_{k}w_{k}}^{c})
=\frac{1}{h_{k} w_{k}} \sum_{i=1}^{h_{k}} \sum_{j=1}^{w_{k}} p_{k-ij}^{c}\quad.
\label{aggrefunc5}
\end{equation}  
\textbf{Training DMIL-WDDS}. The training data is a collection of images $\mathbf{B}_{k}$, $k=1,\ldots,N$ with image-level labels $\mathbf{t}_{k} \in \{0,1\}^{C}$, which is a $C$-dimensional one-hot vector composed by elements $t_{k}^{c}$, $c=1,\dots,C$. We denote by $\mathbb{\psi}(\mathbf{B}_{k}|\mathbf{w})$ our network architecture, mapping an image $\mathbf{B}_{k}$ into a vector of class scores $\mathbf{p}_{k}=[p_k^1,\ldots,p_k^C] \in \mathbb{R}^{C}$, i.e. $\mathbf{p}_{k} = \mathbb{\psi}(\mathbf{B}_{k}|\mathbf{w})$, where $\mathbf{w}$ denotes the weights of our FCN. The following loss function is used to optimise our DMIL-WDDS:
\begin{eqnarray}
\mathcal{L}(\mathbf{w}) &=& \frac{1}{N}\sum_{k=1}^{N} \frac{1}{2}\sum_{c=1}^{C} \big[\mathbb{\psi}^c(B_{k}|\mathbf{w}) - t_{k}^c\big]^2 
 + \frac{\mathrm{\lambda}}{2} ||\mathbf{w}||^2 \nonumber\\
 &=& \frac{1}{2N}\sum_{k=1}^{N} \sum_{c=1}^{C} \big[p_k^c - t_{k}^c\big]^2 
 + \frac{\mathrm{\lambda}}{2} ||\mathbf{w}||^2,
\label{lossfunc}
\end{eqnarray}  
which is aimed at minimizing the mean-square error (MSE) between image-level labels and predictions as well as regularizing the model parameters.
\\\\
\textbf{Identification and localization}. Given one test image $\mathbf{B}_{x}$, we can obtain $C$ spatial score maps $\{p_{x-ij}^c,c=1,\ldots,C\}_{i,j=1}^{h_x,w_x}$ by FCN, which are aggregated into $p_{x}^c$ by MIL framework as showed in Equation~(\ref{aggrefunc3}) (\ref{aggrefunc4}) (\ref{aggrefunc5}). Hence, we make an identification $I_{x}$ for image $\mathbf{B}_{x}$ as follow:
\begin{equation}
I_{x} = \argmax\limits_{c=1,\dots,C} \mathbb{\psi}^c(\mathbf{B}_{x}|\mathbf{w})=\argmax\limits_{c=1,\dots,C} p_{x}^c\quad.
\label{predfunc}
\end{equation}  
Meanwhile, the $C$ spatial score maps go through an up-sampling operation to generate $C$ heat maps each of which owns the same size with image $\mathbf{B}_{x}$. These heat maps are able to roughly delineate the disease areas, \emph{i.e.} the higher heat value in one heat map, the greater possibility of corresponding disease existing. Finally, the BBA operation is performed on the heat map corresponding to label $I_{x}$ to output some bounding boxes locking disease areas (see Section~\ref{sec3.4} for details).
\\\\
\textbf{Real-time design}. Due to the massive storage needs for parameters in deep model and huge computing cost that mobile phone chips can't afford, it is unrealistic to directly embed our deep model into mobile handsets. Therefore, we just implant data acquisition module and interaction module in mobile side, but the decision and calculation module in sever side, including the trained FCN model. The data transmission between mobile and sever is based on HTTP protocol. Given the well-trained FCN model, our DMIL-WDDS allows automatic identification and localization for wheat diseases on-the-fly. Under the condition of China Mobile'4G service, the processing speed can reach almost 1s/image on a Geforce GTX 1080 GPU, which satisfies real-time application.

\begin{table*}[htpb]
\begin{center}
\caption{Detailed architectures of four deep models used in our experiments. The ``LRN'' is Local Response Normalisation \citep{krizhevsky2012imagenet}. The ``stride'' is the convolution stride and the ``pad'' is the spatial padding on feature maps. The nonlinear activation function for all weights layers (except for ``MIL aggregation'') is ReLu.}
\vspace{-0.5em}
\scalebox{0.95}{
\begin{tabular}{lllll}
\hline
  &VGG-CNN-S&VGG-FCN-S&VGG-CNN-VD16&VGG-FCN-VD16 \\ 
\hline
\multirow{4}{*}{Block1} 
& conv-7$\times$7$\times$96 & conv-7$\times$7$\times$96 & conv-3$\times$3$\times$64 & conv-3$\times$3$\times$64 \\
& LRN& LRN & conv-3$\times$3$\times$64 & conv-3$\times$3$\times$64 \\
& stride-2, pad-0 & stride-2, pad-0 & stride-1, pad-1 & stride-1, pad-1 \\
& maxpool-3$\times$3 & maxpool-3$\times$3 &maxpool-2$\times$2 & maxpool-2$\times$2   \\
\hline
\multirow{4}{*}{Block2} 
& conv-5$\times$5$\times$256 & conv-5$\times$5$\times$256 & conv-3$\times$3$\times$128 & conv-3$\times$3$\times$128 \\
& stride-1, pad-1  & stride-1, pad-1 & conv-3$\times$3$\times$128 & conv-3$\times$3$\times$128 \\
& maxpool-2$\times$2  & maxpool-2$\times$2 & stride-1, pad-1 & stride-1, pad-1  \\
&   &  & maxpool-2$\times$2 & maxpool-2$\times$2 \\
\hline
\multirow{5}{*}{Block3} 
& conv-3$\times$3$\times$512 & conv-3$\times$3$\times$512 & conv-3$\times$3$\times$256 & conv-3$\times$3$\times$256 \\
& stride-1, pad-1  & stride-1, pad-1& conv-3$\times$3$\times$256 & conv-3$\times$3$\times$256 \\
& &  & conv-3$\times$3$\times$256 & conv-3$\times$3$\times$256 \\
& &  &stride-1, pad-1 & stride-1, pad-1  \\
& &  & maxpool-2$\times$2 & maxpool-2$\times$2 \\
\hline
\multirow{5}{*}{Block4} 
& conv-3$\times$3$\times$512 & conv-3$\times$3$\times$512 & conv-3$\times$3$\times$512 & conv-3$\times$3$\times$512 \\
& stride-1, pad-1  & stride-1, pad-1 & conv-3$\times$3$\times$512 & conv-3$\times$3$\times$512 \\
& &  & conv-3$\times$3$\times$512 & conv-3$\times$3$\times$512 \\
& &  &stride-1, pad-1 & stride-1, pad-1  \\
& &  & maxpool-2$\times$2 & maxpool-2$\times$2 \\
\hline
\multirow{5}{*}{Block5} 
& conv-3$\times$3$\times$512 & conv-3$\times$3$\times$512 & conv-3$\times$3$\times$512 & conv-3$\times$3$\times$512 \\
& stride-1, pad-1  & stride-1, pad-1& conv-3$\times$3$\times$512 & conv-3$\times$3$\times$512 \\
& maxpool-3$\times$3& maxpool-3$\times$3 & conv-3$\times$3$\times$512 & conv-3$\times$3$\times$512 \\
& &  &stride-1, pad-1 & stride-1, pad-1  \\
& &  & maxpool-2$\times$2 & maxpool-2$\times$2 \\
\hline
\multirow{2}{*}{Block6}
& fc-1024 & conv 6$\times$6$\times$1024 & fc-1024 & conv-7$\times$7$\times$1024\\
&  & stride-1, pad-0 &  & stride-1, pad-0\\
\hline
\multirow{2}{*}{Block7}
& fc-1024 & conv 1$\times$1$\times$1024 & fc-1024 & conv 1$\times$1$\times$1024\\
&  & stride-1, pad-0 &  & stride-1, pad-0\\
\hline
Block8 &softmax-7 & MIL aggregation & softmax-7 & MIL aggregation\\
\hline
\end{tabular}}
\label{Tab2}
\end{center}
\end{table*}

\section{Experiments}
\label{sec4}
We verify the effectiveness of proposed DMIL-WDDS upon WDD2017 by performing some experiments in which several models are built to estimate the test accuracy and show the localization results. Below we give details of our experiments.

\subsection{Models}
\label{sec4.1}
We consider the modified VGG-CNN-S and VGG-CNN-VD16 as our baseline model. Based on the two conventional CNN models, we develop the VGG-FCN-S and VGG-FCN-VD16 as our basic model of DMIL-WDDS framework according to the paradigm mentioned in Section~\ref{sec3.3}. Detailed architectures for the four models are showed in Table~\ref{Tab2}. Note that the VGG-CNN-S and VGG-FCN-S have the same amount of parameters, as well as for VGG-CNN-VD16 and VGG-FCN-VD16. The MIL aggregation has three forms: Soft-agg, Max-agg and Avg-agg, which correspond to Equation~(\ref{aggrefunc3}) (\ref{aggrefunc4}) (\ref{aggrefunc5}) respectively. The constant $\alpha$ is set as 2.5 in our experiments.

\subsection{Dataset}
\label{sec4.2}
We evaluate our DMIL-WDDS on the collected in-field wheat crop dataset WDD2017. For conventional CNN architectures, \emph{i.e.} VGG-CNN-S and VGG-CNN-VD16, we resize all RGB images in WDD2017 into the size of $224\times224\times3$. But for DMIL-WDDS, \emph{i.e.} VGG-FCN-S and VGG-FCN-VD16, we resize all RGB images into the size of $832\times832\times3$. Moreover, we uniformly split the WDD2017 into 5 folds for cross-validation (detailed split is showed in Table~\ref{Tab1}).

\subsection{Training strategy}
\label{sec4.3}
The Nesterov Momentum SGD is used as our optimizer with momentum 0.9. For conventional CNN architectures, the VGG-CNN-S and VGG-CNN-VD16 are trained for 60 epochs with a batch size of 45 examples and initial learning rate of 0.0001, and the learning rate is divided by 10 per 10 epochs. For DMIL-WDDS, VGG-FCN-S and VGG-FCN-VD16 are trained for 20 epochs with a batch size of 2 examples and intial learning rate of 0.00005, and the learning rate is divided by 5 per 5 epochs. In addition, for the four models we fine tune the parameters of the first 5 blocks which is pre-trained on the ImageNet ILSVRC2012, but start from scratch for the last 3 transformed blocks. 

\subsection{Accuracy estimation}
\label{sec4.4}
For each deep model, we report our test accuracy for every wheat disease as well as total accuracy in an average manner over 5-fold cross-validation. The hyperparameters for the four deep models are chosen such as to maximize the recognition accuracy over a random validation set from training folds.

\subsection{Implementation}
\label{sec4.5}
 Our codes\footnote{Our codes will be released if this paper is accepted.} are written in Python and the deep models are implemented with Theano \citep{al2016theano}. We run our codes on a Geforce GTX 1080 GPU.

\section{Results and discussion}
\label{sec5}

\subsection{Wheat disease identification}
\label{sec5.1}

\begin{table*}[tpb]
\begin{center}
\caption{Accuracy (\%) for every class as well as total in WDD2017 for conventional CNN architectures and DMIL-WDDS models (mean $\pm$ std ). The Max-agg, Avg-agg and Soft-agg correspond to the three different aggregated function of MIL framework in DMIL-WDDS. The bold value is the best performance for corresponding deep model along each category.}
\vspace{-0.5em}
\scalebox{0.75}{
\begin{tabular}{lllllclll}
\hline
\multirow{2}{*}{Model} & \multirow{2}{*}{VGG-CNN-S} & \multicolumn{3}{c}{VGG-FCN-S} & \multirow{2}{*}{VGG-CNN-VD16} & \multicolumn{3}{c}{VGG-FCN-VD16} \\
\cline{3-5}  \cline{7-9}  
 & & Max-agg& Avg-agg&Soft-agg& & Max-agg& Avg-agg&Soft-agg\\ 
\hline
Powdery Mildew(\%) &  2.00 $\pm$ 1.94& 50.29 $\pm$ 8.20& 48.86 $\pm$ 3.66& \textbf{66.57 $\pm$ 3.90} & 66.57 $\pm$ 5.00& 84.86 $\pm$ 2.94& 79.43 $\pm$ 2.32& \textbf{89.14 $\pm$ 7.43}\\
Smut (\%) &  87.28 $\pm$ 2.27& 91.69 $\pm$ 3.15& 93.75 $\pm$ 1.51& \textbf{96.84 $\pm$ 1.54}& 96.98 $\pm$ 1.99& 98.08 $\pm$ 0.80& 98.28 $\pm$ 0.48& \textbf{99.18 $\pm$ 0.46}\\
Black Chaff (\%) &  89.91 $\pm$ 3.35& 91.79 $\pm$ 3.09& 93.51 $\pm$ 2.27& \textbf{93.68 $\pm$ 1.99}& \textbf{96.24 $\pm$ 1.76} & 95.05 $\pm$ 1.74& 95.56 $\pm$ 1.98& 95.90 $\pm$ 1.57\\
Stripe Rust (\%) &  59.72 $\pm$ 3.60& 88.66 $\pm$ 1.03& 90.31 $\pm$ 1.63& \textbf{95.50 $\pm$ 0.55}& 89.91 $\pm$ 1.91& 95.39 $\pm$ 1.04& 95.78 $\pm$ 1.29& \textbf{96.75 $\pm$ 0.53}\\
Leaf Blotch (\%) &  82.44 $\pm$ 3.05& 92.75 $\pm$ 1.36& \textbf{96.86 $\pm$ 0.44}& 96.42 $\pm$ 0.49& 95.85 $\pm$ 1.02& 97.35 $\pm$ 0.32& 98.49 $\pm$ 0.27& \textbf{98.66 $\pm$ 0.49}\\
Leaf Rust (\%) &  62.70 $\pm$ 6.11& 93.87 $\pm$ 1.77& 95.41 $\pm$ 1.35& \textbf{96.67 $\pm$ 1.29}& 92.43 $\pm$ 1.12& 98.47 $\pm$ 0.78& 98.20 $\pm$ 1.10& \textbf{98.74 $\pm$ 0.44}\\
Healthy Wheat (\%) &  76.77 $\pm$ 4.28& 93.16 $\pm$ 1.81& 96.51 $\pm$ 0.94& \textbf{96.91 $\pm$ 0.71}& 95.07 $\pm$ 2.49& 98.22 $\pm$ 1.22& 98.16 $\pm$ 0.97& \textbf{99.28 $\pm$ 0.73}\\
\hline
Total (\%) & 73.00 $\pm$ 1.07& 90.34 $\pm$ 0.93 & 92.86 $\pm$ 0.78 & \textbf{95.12 $\pm$ 0.59}& 93.27 $\pm$ 0.72& 96.75 $\pm$ 0.23& 96.94 $\pm$ 0.40& \textbf{97.95 $\pm$ 0.36}\\
\hline
\end{tabular}}
\label{Tab3}
\end{center}
\end{table*}

Table~\ref{Tab3} presents the final class-wise accuracy and total accuracy over 5-fold cross-validation for different deep model configurations. To compare our DMIL-WDDS framework with conventional CNN architectures, we firstly observe the results of VGG-CNN-S and VGG-FCN-S. Note that the VGG-FCN-S with three aggregated functions (Max-agg, Avg-agg and Soft-agg) significantly and consistently outperforms VGG-CNN-S on all classes, improving the total accuracy by 17.34\%, 19.86\% and 22.12\% respectively. Also we can see the homologous advance on accuracy from the results of VGG-CNN-VD16 and VGG-FCN-VD16. The VGG-FCN-VD16 with three aggregated functions outperforms VGG-CNN-VD16 in all classes except for ``Black Chaff'', improving the total accuracy by 3.48\%, 3.67\% and 4.68\% respectively. In brief, the proposed DMIL-WDDS framework surpasses conventional CNN-based architectures on identification task under the same amount of model parameters.

 \begin{figure}[htbp]
\begin{center}
\includegraphics[width=0.48\linewidth]{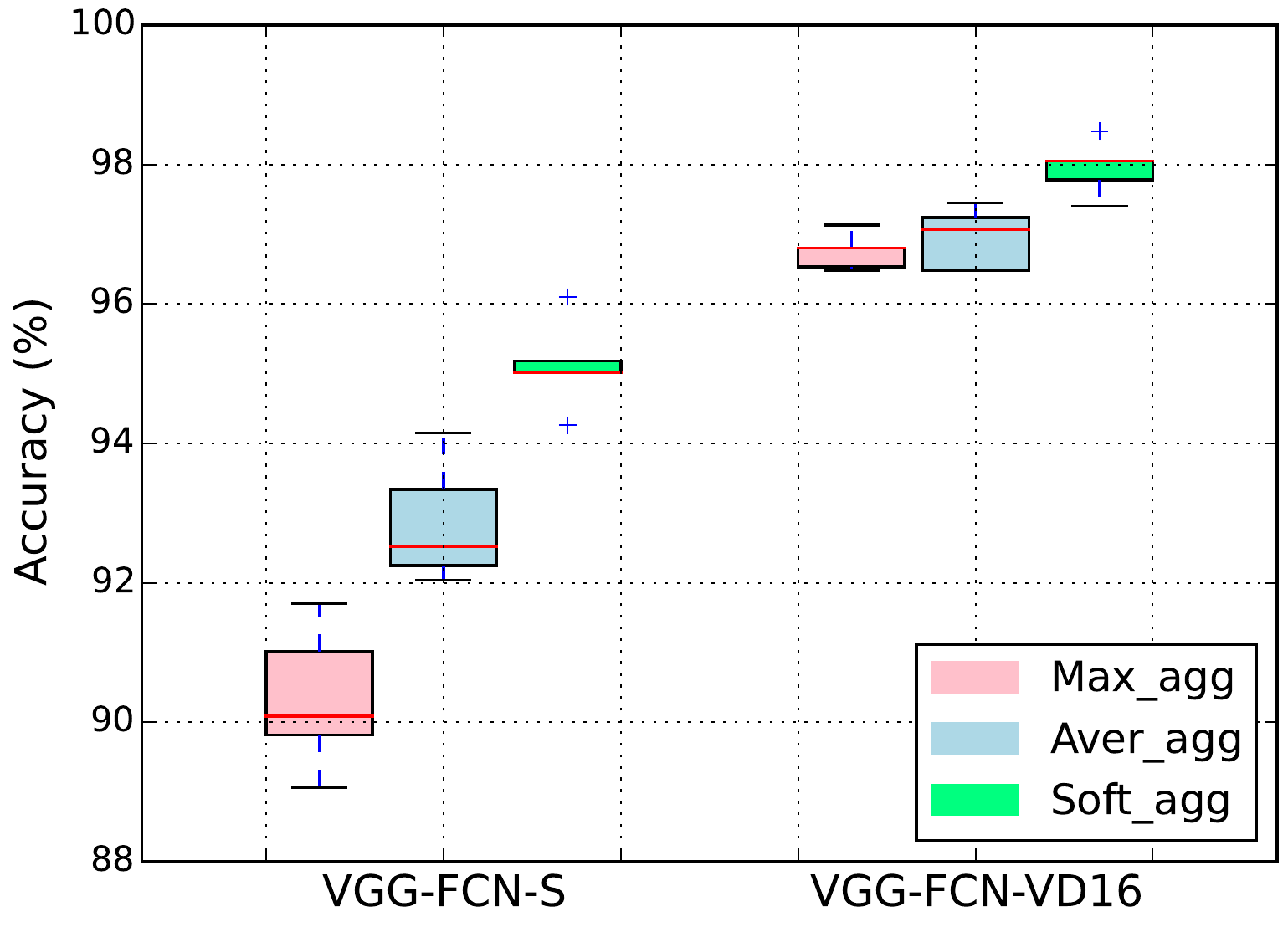}
\end{center}
\vspace{-1.5em}
  \caption{Total accuracy comparison between DMIL-WDDS models with different aggregated MIL function. The performances of all DMIL-WDDS models with different aggregated functions over 5-fold cross-validation are showed as boxplots. Best viewed in color.}
\label{boxfig} 
\end{figure}

Figure~\ref{boxfig} shows the comparison of total accuracies from different aggregated MIL functions for DMIL-WDDS models. It can be seen that Soft-agg achieve the best performance for both DMIL-WDDS models, and the Avg-agg comes second, while the Max-agg is relatively worst. These results suggest that it is beneficial for in-field wheat disease identification to take into account for all respective fields in a weighted manner.

\begin{figure}[htbp]
\begin{center}
\includegraphics[width=0.9\linewidth]{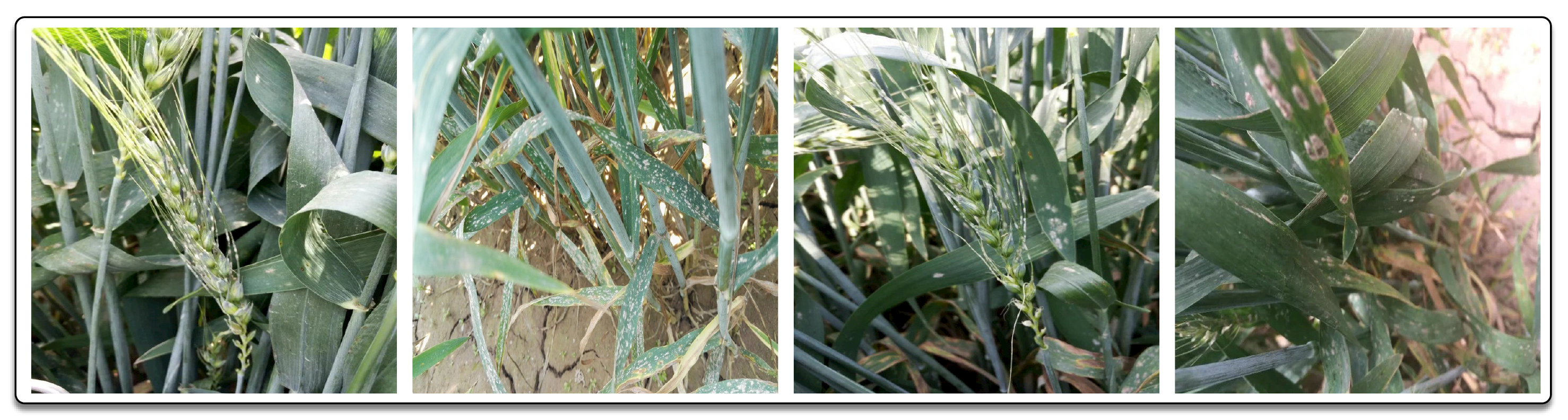}
\end{center}
\vspace{-2.0em}
  \caption{Examples of wheat ``Powdery Mildew''. Its visual symptoms are tiny white spots in leaves or ears of wheat. Best viewed in color.}
\label{Powdery_Mildew} 
\end{figure}

Besides, another two amazing observations can be derived in Table~\ref{Tab3}. The first one is that the shallow model in DMIL-WDDS framework may be superior to deep model in conventional CNN architecture for in-field wheat disease identification tasks, for example, the 95.12\% of VGG-FCN-S exceeds 93.27\% of VGG-CNN-VD16. The second one is that the DMIL-WDDS framework can effectively deal with some samples that conventional CNN is unable to deal with. For instance, ``Powdery Mildew'' is one kind of wheat disease that is distinguishable by CNN, as showed in Figure~\ref{Powdery_Mildew}, because its visual symptoms are tiny white spots which are easily filtered out by pooling operations of CNN. VGG-CNN-S suffer a heavy breakdown on the identification of ``Powdery Mildew'', whose class-wise accuracy is just 2.00\%, but VGG-FCN-S with Soft-agg can achieve a class-wise accuracy of 66.57\% on this class, which is a significant performance boost. The two observations can be explained by that DMIL-WDDS frameworks focus on the fine feature extraction for local areas in whole image.

\subsection{Contrast test and feature visualization}
\label{sec5.2}

\begin{figure}[tbp]
\centering
\subfloat[Example image suffering from ``Stripe Rust''.]{
 \includegraphics[width=0.485\linewidth]{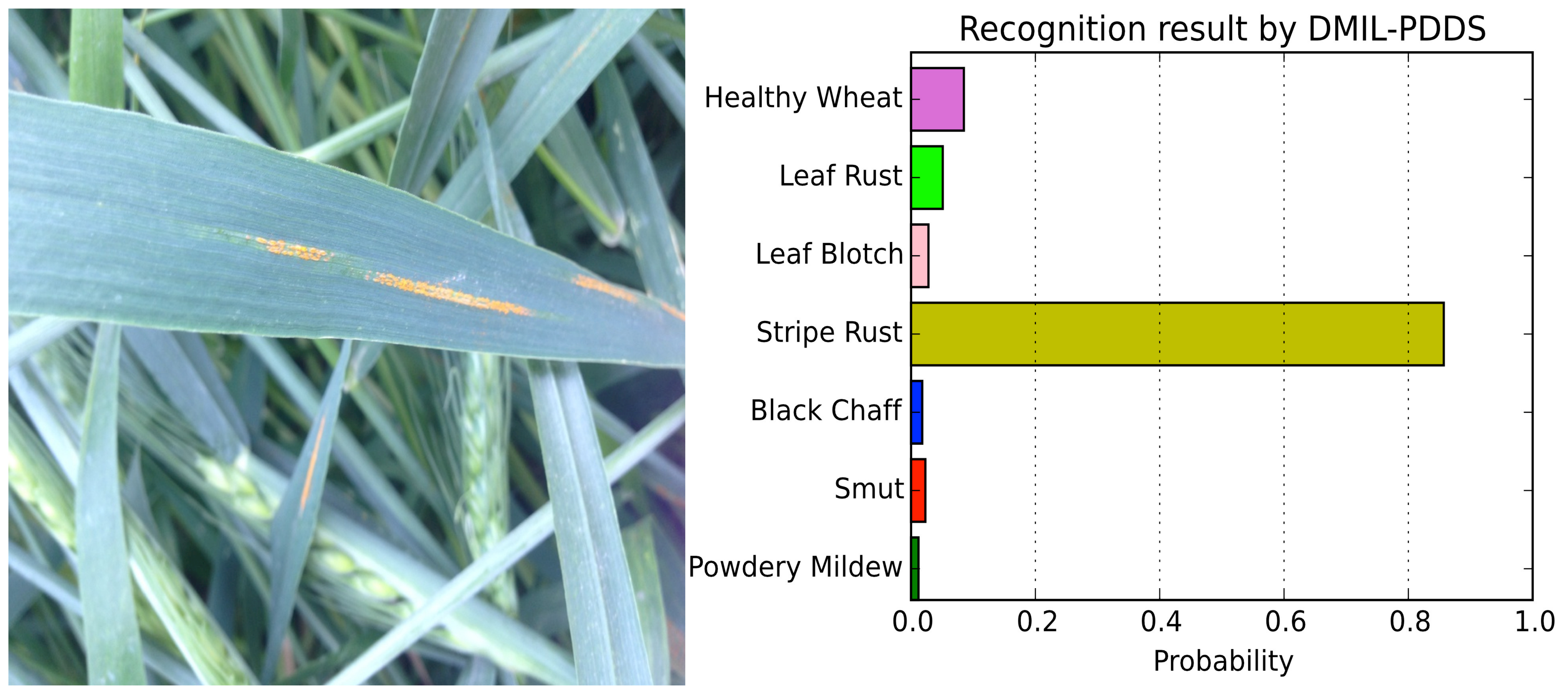}
}
\hspace{-0.6em}
\subfloat[Diseased areas of ``Stripe Rust'' are shielded.]{
 \includegraphics[width=0.485\linewidth]{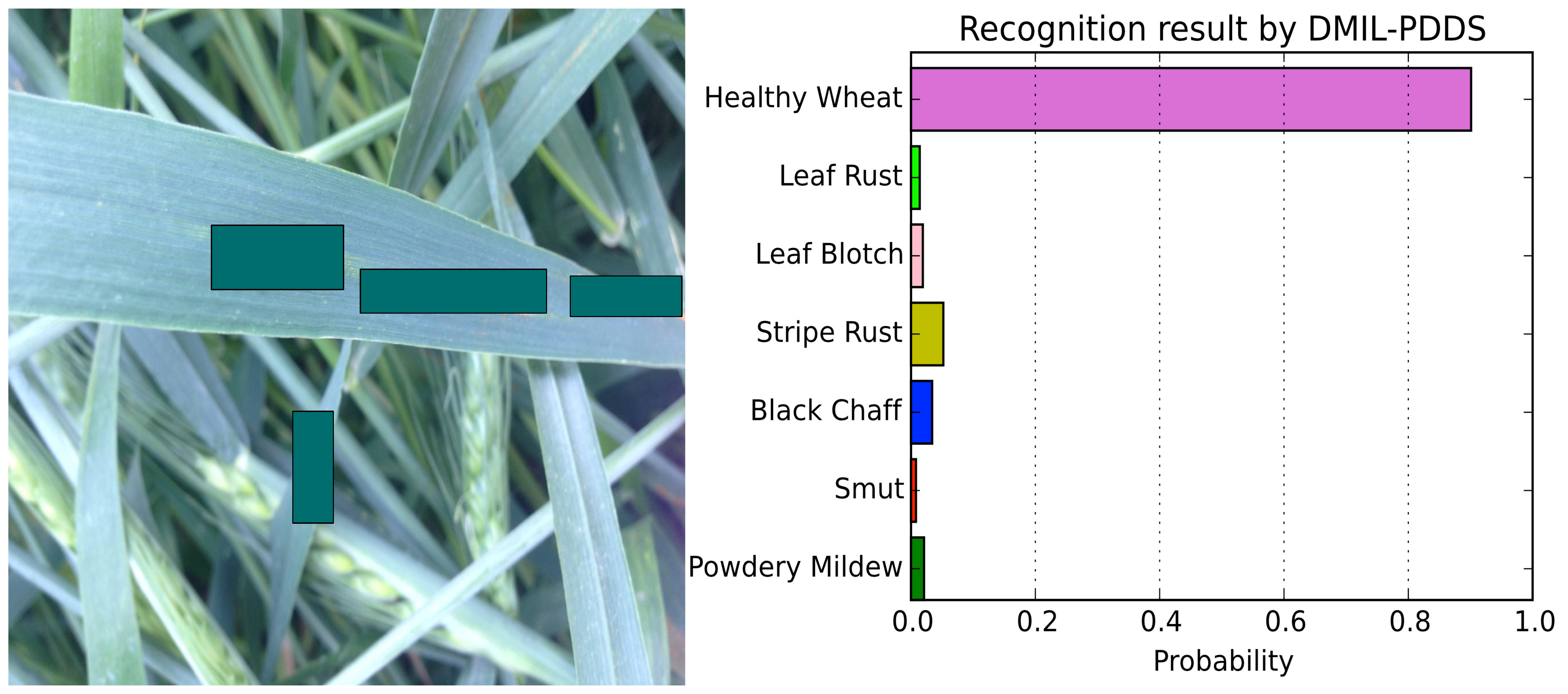}
}
\vspace{-0.5em}
\caption{Comparison of recognition results for raw image and processed image (shielding diseased areas).}
\label{example}
\end{figure}

\begin{figure*}[tbp]
\centering
\subfloat[feature maps outputted by Block 1.]{
 \includegraphics[width=0.4\linewidth]{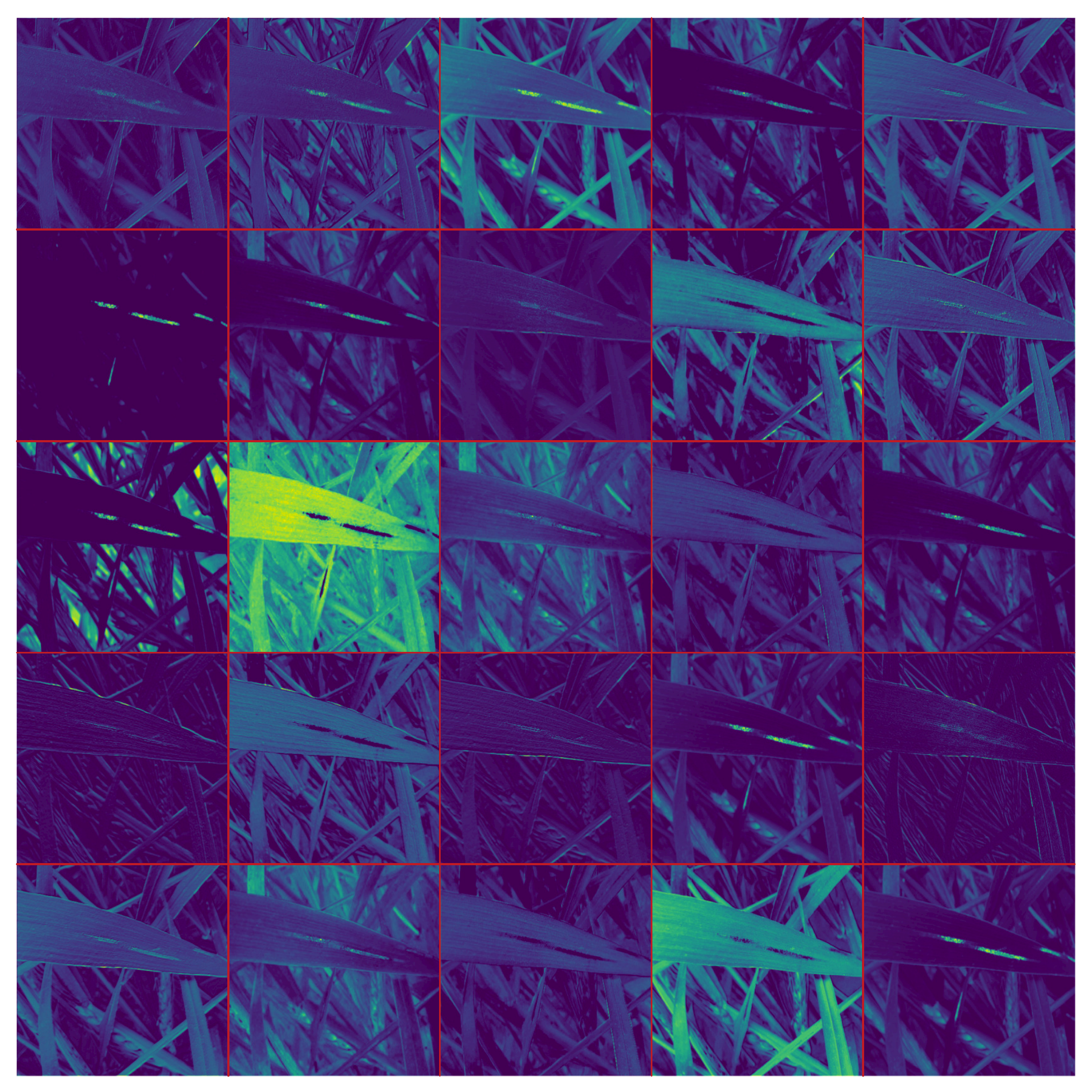}
}
\hspace{1em}
\subfloat[feature maps outputted by Block 2.]{
 \includegraphics[width=0.4\linewidth]{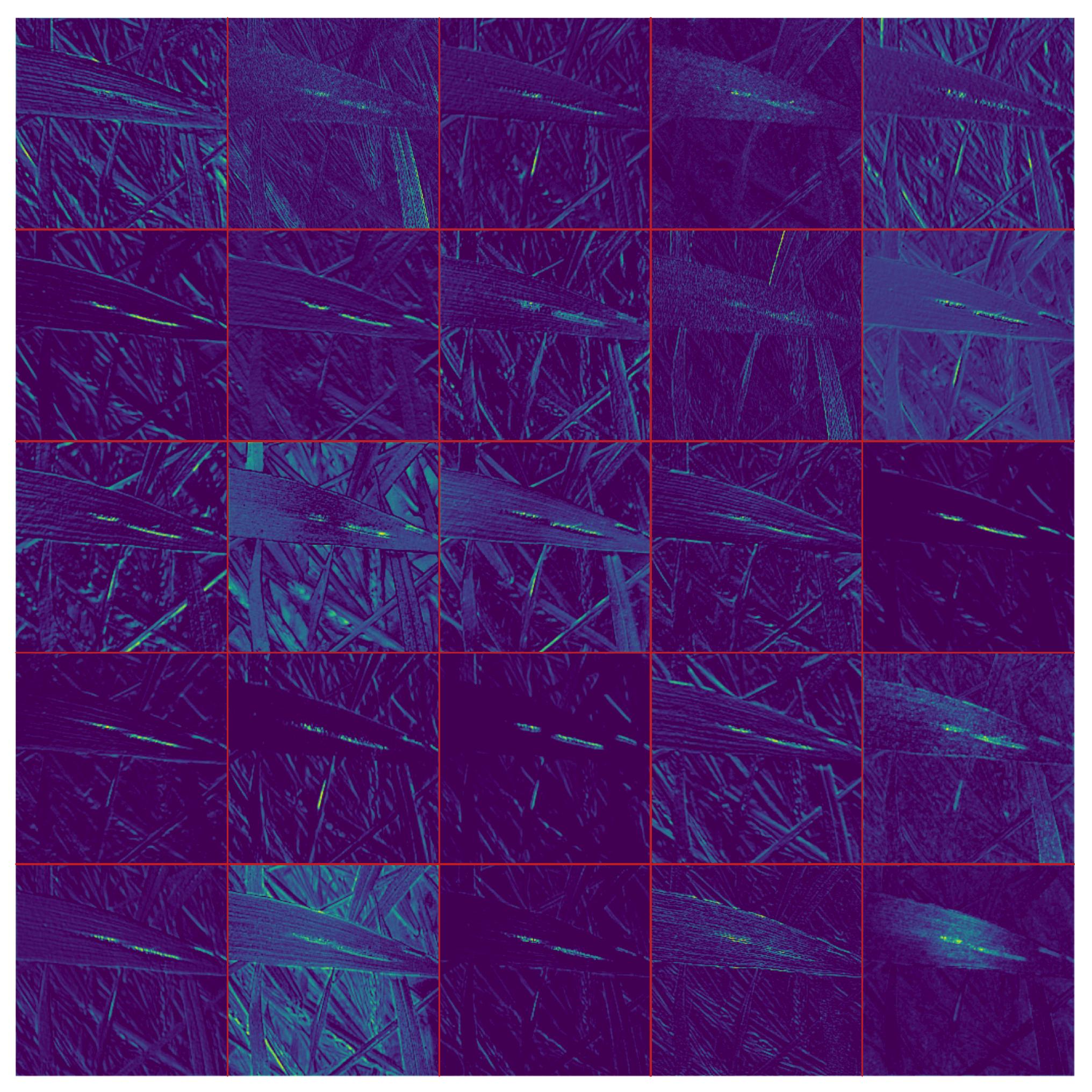}
}
\hspace{1em}
\subfloat[feature maps outputted by Block 3.]{
 \includegraphics[width=0.4\linewidth]{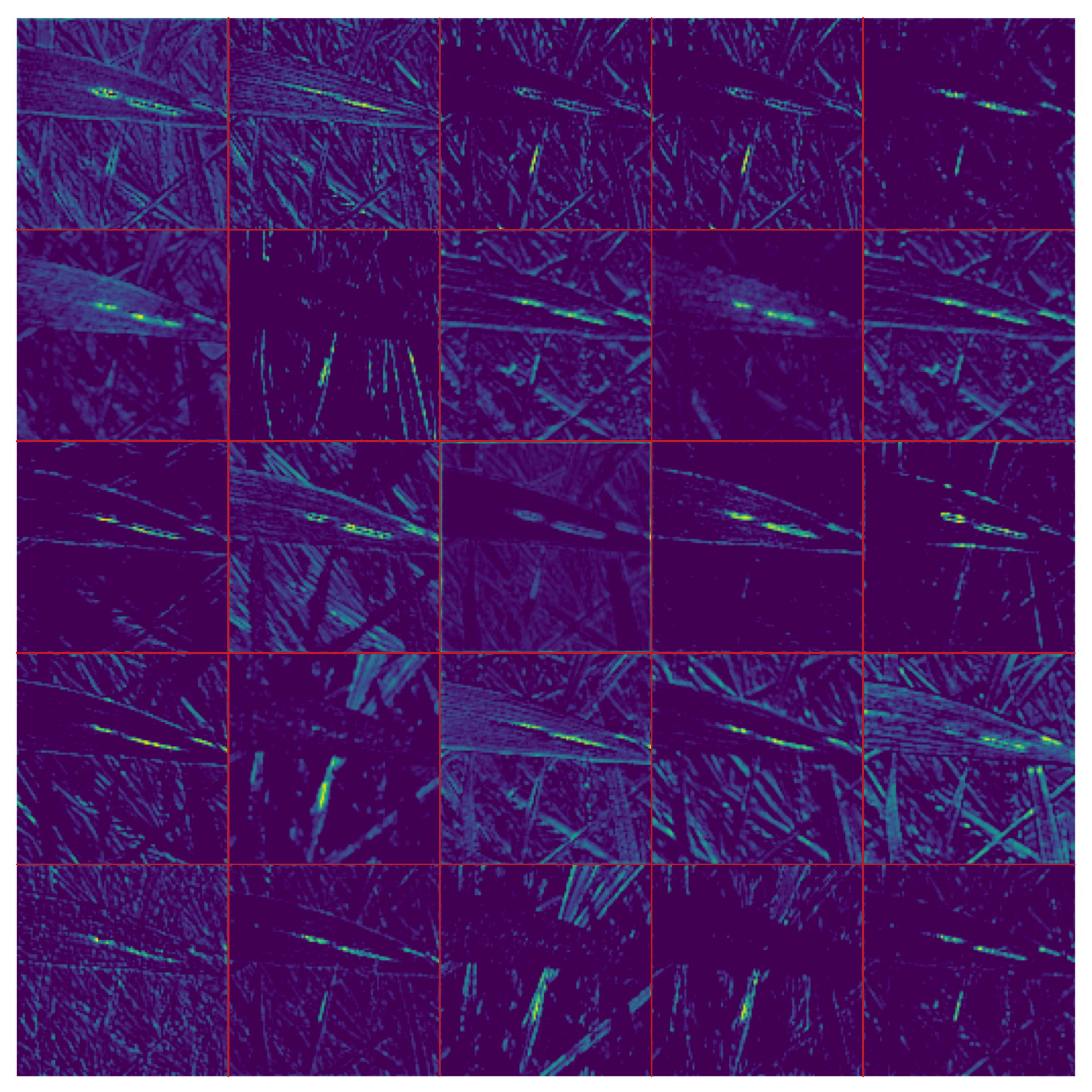}
}
\hspace{1em}
\subfloat[feature maps outputted by Block 4.]{
 \includegraphics[width=0.4\linewidth]{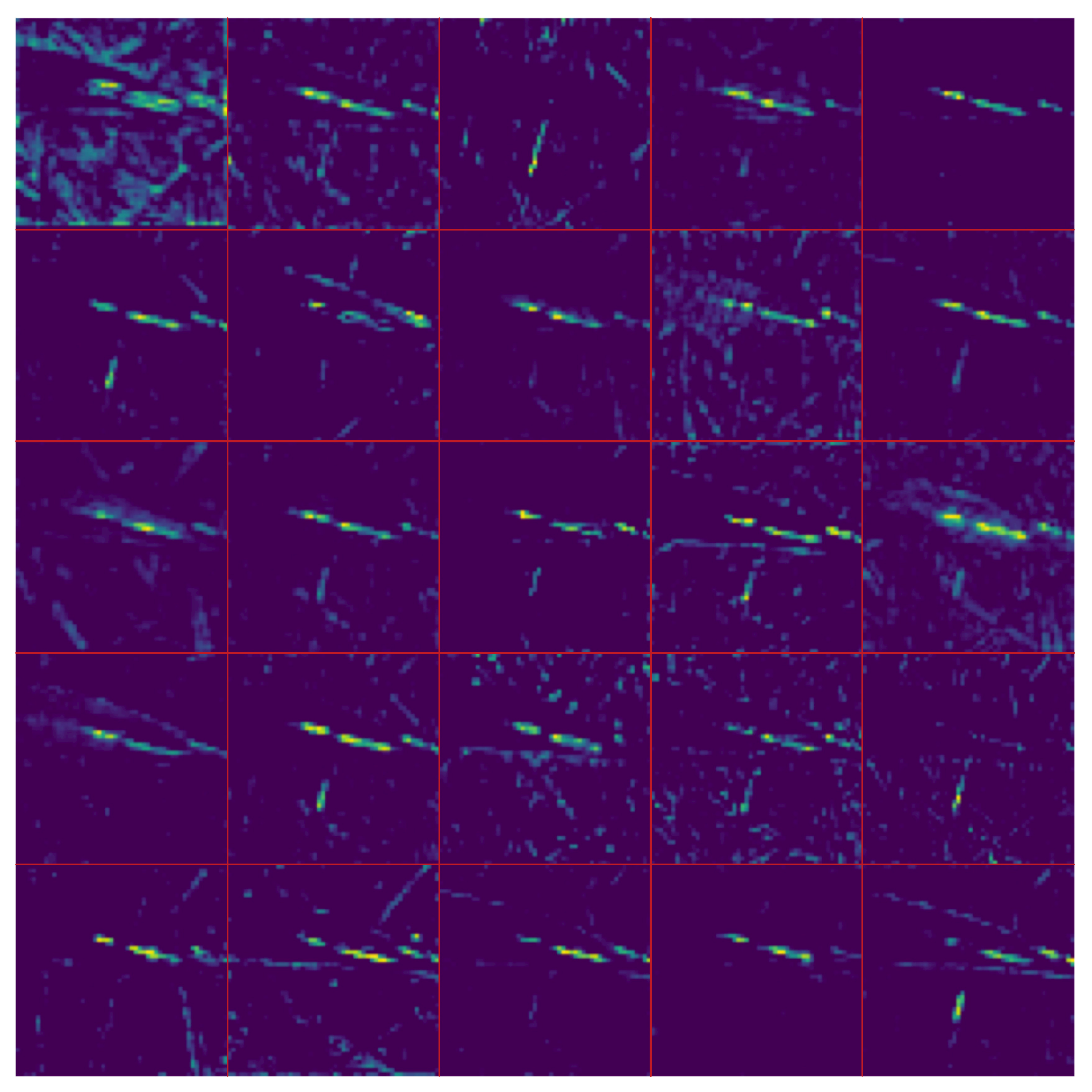}
}
\vspace{-0.5em}
\caption{Visualization of features for Figure~\ref{example}(a) in trained DMIL-WDDS based on VGG-FCN-VD16 model. Note that only the first 25 feature maps after Block 1-4 have been illustrated due to space limitations.
}
\label{fea_block}
\end{figure*}

\begin{figure*}[htbp]
\begin{center}
\includegraphics[width=0.92\linewidth]{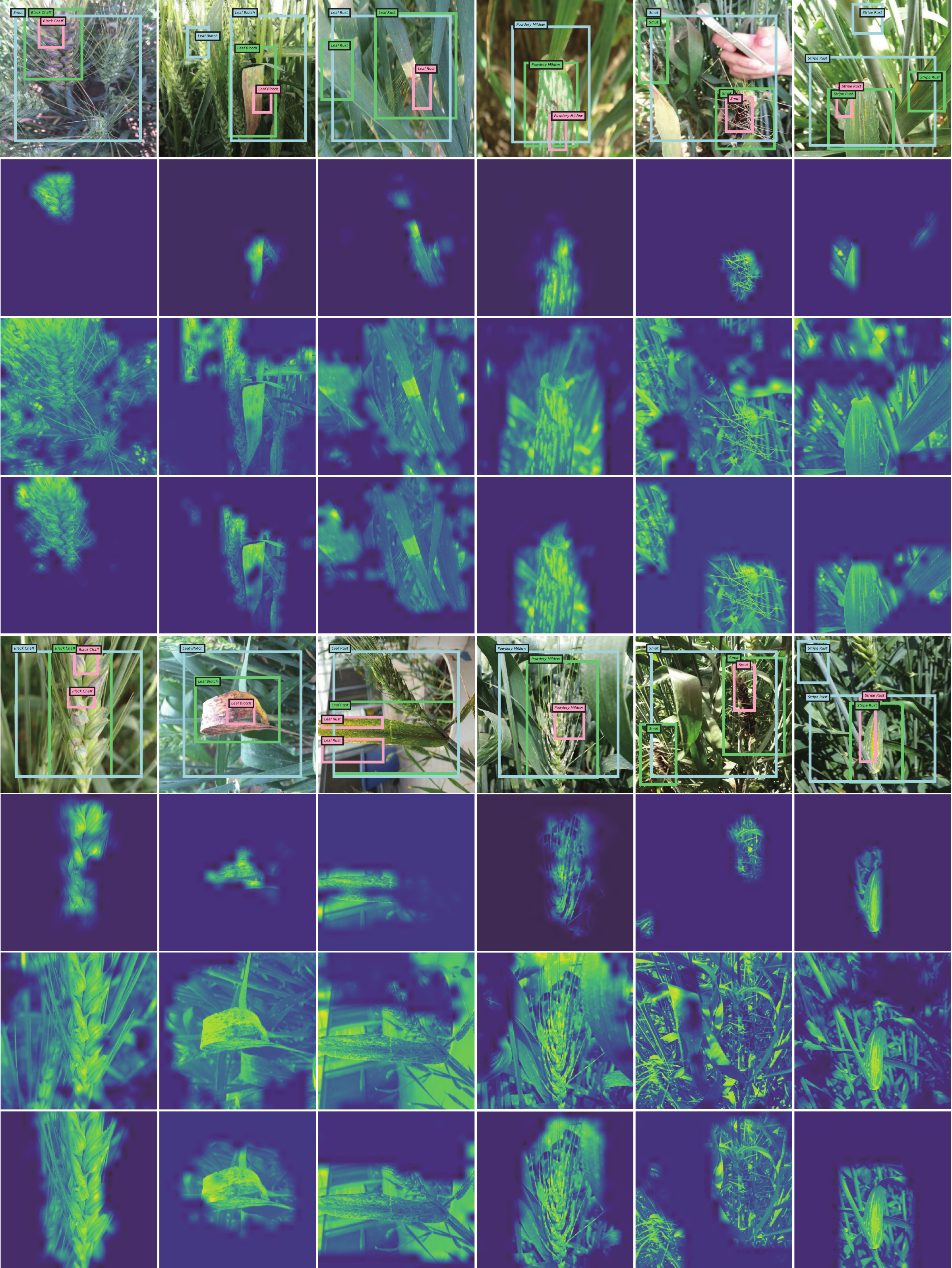}
\end{center}
\vspace{-1.0em}
  \caption{Examples of localization for wheat disease areas by DMIL-WDDS (VGG-FCN-VD16). Six columns of images from left to right are respectively ``Black Chaff'', ``Leaf Blotch'', ``Leaf Rust'', ``Powdery Mildew'', ``Smut'' and ``Leaf Rust''. Pink, lightblue and green boxes in raw images indicate the localization results by VGG-FCN-VD16 with Max-agg, Avg-agg and Soft-agg respectively. Three heat maps below raw images, which are generated by masking spatial score maps on B channel in RGB images, figure out the disease areas inferred by VGG-FCN-VD16 with Max-agg, Avg-agg and Soft-agg respectively. Best viewed in color.}
\label{localization} 
\end{figure*}

In order to better explain what our system is actually learning, the contrast tests have been carried out. For example, one raw image suffering from ``Stripe Rust'' and its processed version (by shielding diseased areas) are prepared for diagnosis by DMIL-WDDS (using VGG-FCN-VD16 model). Figure~\ref{example} demonstrates the comparison of recognition results for the pair of images. As shown in Figure~\ref{example}(a), the raw image is correctly classified into ``Stripe Rust'' by DMIL-WDDS. However, as its diseased areas are shielded by some blocks, the recognition result turns to ``Healthy Wheat'' as shown in Figure~\ref{example}(b). This phenomenon indicates that the DMIL-WDDS is indeed sensitive to disease areas rather than other parts of the wheat.

For further investigations of the learning capacity of our DMIL-WDDS, we visualize the features of the raw ``Stripe Rust'' image (as shown in Figure~\ref{example}(a)) in Figure~\ref{fea_block}, where (a) to (d) reveal the feature maps outputted by Block 1 to 4 of VGG-FCN-VD16, respectively. Note that a pixel point in each feature map is a rectified activation, and the brighter point represents the greater activation value. As we can see, starting from the Block 1 where features of ``Stripe Rust'' go from individual pixels to light bright contours, to the Block 4 where learned features of the four diseased areas are obvious and distinct, our DMIL-WDDS based on VGG-FCN-VD16 achieves a progressive learning process for fine characteristics of disease. This feature visualization is a good demonstration of what our DMIL-WDDS is learning.

\subsection{Localization for disease areas}
\label{sec5.3}
Figure~\ref{localization} presents some examples of localization for wheat disease areas by VGG-FCN-VD16 with three different aggregated functions, namely Max-agg, Avg-agg and Soft-agg. From these observations we can summary that Max-agg results in a too partial localization while Avg-agg results in a too general localization for disease areas. Comparatively, Soft-agg lead to a more precise localization for striking disease areas, which can be explained by that the Soft-agg is a trade-off between the Max-agg and Avg-agg in MIL framework, and we believe that the localization can be refined if given a more suitable hyperparameter $\alpha$ in Equation~(\ref{aggrefunc3}) for Soft-agg. 

Despite the in-field images with intractable challenges mentioned in Section~\ref{sec1}, such as complex image backgrounds or multiple disease areas existing in one image, our DMIL-WDDS can still precisely locate the positions of disease areas, as showed in Figure~\ref{localization}. For example, the two in-field images in the fifth column contains two wheat smut areas as well as people's hands or illumination changes, but our DMIL-WDDS based on VGG-FCN-VD16 with Soft-agg still make correct class-level predictions as well as performing accurate localization for all smut areas (green boxes).

\subsection{Extensibility and future prospects}
\label{sec5.4}
Although the DMIL-WDDS is a task-specific diseases diagnosis system aimed at wheat, the proposed framework is general and suitable for other crops or vegetables, such as corn, tomato, pepper etc. Moreover, the deep architecture of our framework can be adapted or modified to better fit the task data. When our system needs to be extended to some different crops sharing the same diseases with  current crop, we just need relative few data of the new crops about the shared diseases to fine tune our framework to recognize the same diseases but keep the structure of deep architecture unchanged. When our system needs to be extended to some new diseases of current crop, all we need to do is to add corresponding channels in deep architecture and fine tune it based on the new data of added disease classes.

In the near future, we will concentrate on developing a more robust and powerful framework based on current system to fit the mixed cases of multiple types of diseases or multiple crops, which are very difficult challenges for crop disease diagnosis.

\section{Conclusion}
\label{sec6}
In this paper we propose an novel wheat disease diagnosis framework based on deep multiple instance learning, namely DMIL-WDDS, which aims to deal with in-field wheat images without any technical preprocessing. Only with the supply of image-level labelled training data, our system realizes an integration of identification for wheat diseases and localization for disease areas. We exploit four different deep models to perform wheat disease recognition on the newly collected in-field dataset WDD2017. Experimental results on WDD2017 show that the proposed framework provides significant improvements over conventional CNN architectures under the same amount of parameters in deep model. Even based on the shallow model (like VGG-FCN-S), the proposed framework surprisingly makes a better recognition performance than the deep conventional CNN model (like VGG-CNN-VD16). Moreover, the experimental comparison between different aggregated functions in MIL framework indicates that Softmax aggregation is a superior choice for DMIL-WDDS to improve the recognition accuracy. The localization results validate that our DMIL-WDDS with Softmax aggregation can lead to more accurate localization for disease areas than another two aggregated functions. Besides, the proposed DMIL-WDDS has been designed into a mobile app to provide support for agricultural disease diagnosis.

\bibliographystyle{elsarticle-harv}
\bibliography{ref}

\end{document}